\documentclass{article}

\usepackage[final]{neurips_2024}

\usepackage{amsmath}
\usepackage{caption}
\usepackage[utf8]{inputenc} % allow utf-8 input
\usepackage[T1]{fontenc}    % use 8-bit T1 fonts
\usepackage{hyperref}       % hyperlinks
\usepackage{url}            % simple URL typesetting
\usepackage{booktabs}       % professional-quality tables
\usepackage{amsfonts}       % blackboard math symbols
\usepackage{nicefrac}       % compact symbols for 1/2, etc.
\usepackage{microtype}      % microtypography
\usepackage{xcolor}         % colors
\usepackage{listings}
\usepackage{algpseudocode}
\usepackage{algorithm}
\usepackage{hyperref}
\usepackage{float}
\usepackage{bbding}
\usepackage{graphicx}
\usepackage{makecell}
\usepackage{multicol,multirow}
\hypersetup{
    colorlinks=true,
    linkcolor=red,
    anchorcolor=green,
    citecolor=blue}

\makeatletter
\AtBeginDocument{
\let\@oldmaketitle\@maketitle
\renewcommand{\@maketitle}{\@oldmaketitle
  \includegraphics[width=\linewidth]{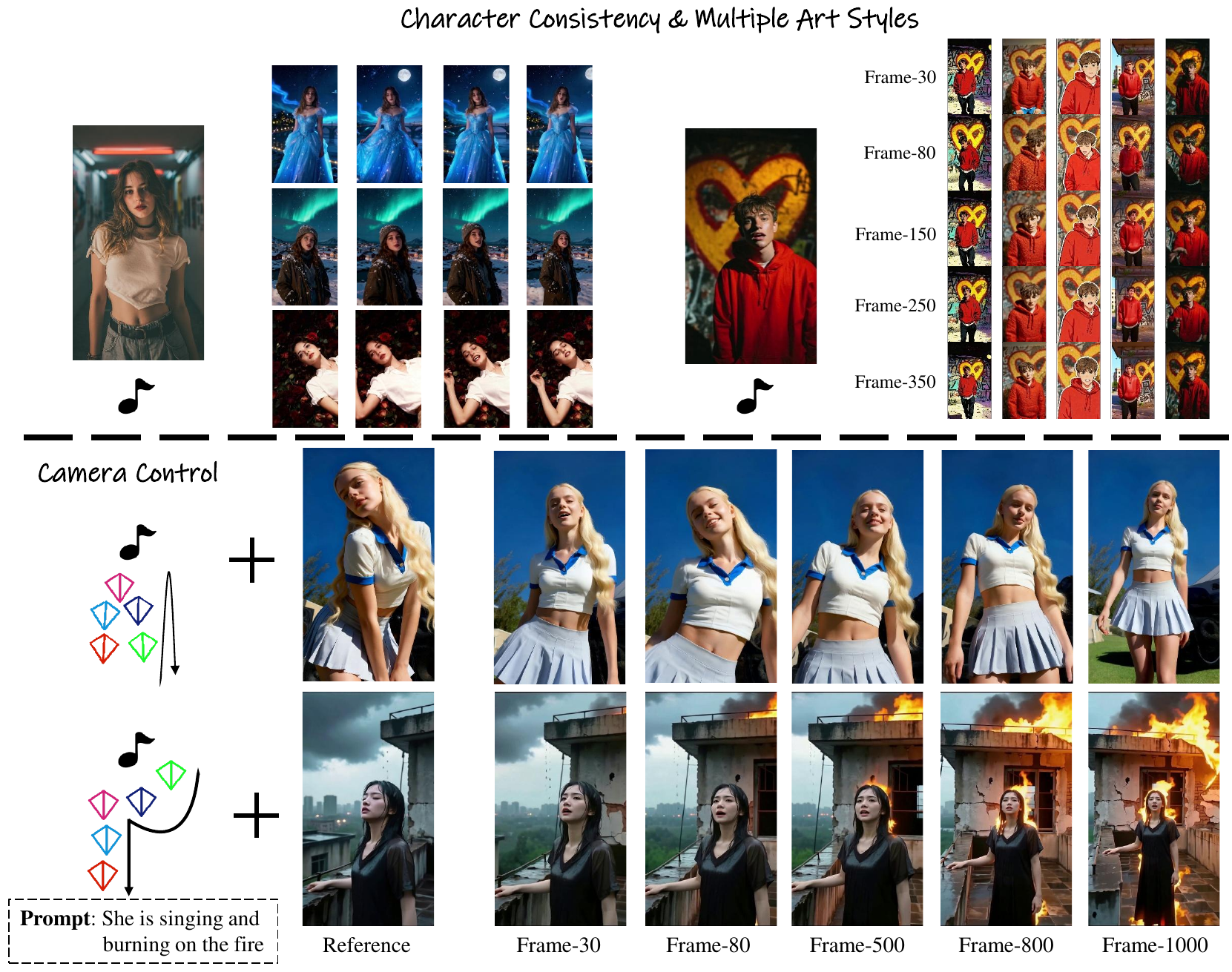}
  \captionof{figure}{
    Conditioned on a portrait image, text, and music input, YingVideo-MV can generate and edit portraits with high identity consistency, expressive facial features, natural body dynamics, and camera movement. The results demonstrate vivid emotions, rich camera movements, and precise lip-syncing, while also generating different artistic styles.
  }
  \label{fig:teaser}
  \vspace{17pt}
 }
}
\makeatother

\title{YingVideo-MV: Music-Driven Multi-Stage Video Generation}

\author{%
  Jiahui Chen \quad
  Weida Wang \quad
  Runhua Shi \quad 
  Huan Yang \quad \\
  \textbf{Chaofan Ding} \quad 
 \textbf{Zihao Chen} \quad \\
 AI Lab, GiantNetwork\\
  \texttt{Shanghai, China} \\
}
\begin{document}

\maketitle

\begin{abstract}
While diffusion model for audio-driven avatar video generation have achieved notable process in synthesizing long sequences with natural audio-visual synchronization and identity consistency, the generation of music-performance videos with camera motions remains largely unexplored. We present YingVideo-MV, the first cascaded framework for music-driven long-video generation. Our approach integrates audio semantic analysis, an interpretable shot planning module (MV-Director), temporal-aware diffusion Transformer architectures, and long-sequence consistency modeling to enable automatic synthesis of high-quality music performance videos from audio signals. We construct a large-scale Music-in-the-Wild Dataset by collecting web data to support the achievement of diverse, high-quality results. Observing that existing long-video generation methods lack explicit camera motion control, we introduce a camera adapter module that embeds camera poses into latent noise. To enhance continulity between clips during long-sequence inference, we further propose a time-aware dynamic window range strategy that adaptively adjust denoising ranges based on audio embedding. Comprehensive benchmark tests demonstrate that YingVideo-MV achieves outstanding performance in generating coherent and expressive music videos, and enables precise music-motion-camera synchronization. More videos are available in our project page: \href{https://giantailab.github.io/YingVideo-MV/}{https://giantailab.github.io/YingVideo-MV/}.
\end{abstract}

\section{Introduction}

Music-performing avatars have demonstrated significant research and application value across diverse visual media creation domains, including cinema, music videos (MV), vlogs, and advertisements. These models~\cite{yang2025infinitetalk, ding2025kling, tu2025stableavatar} achieve temporally coherent facial expressions, lip movements~\cite{fei2025skyreels}, and body poses~\cite{wang2025fantasytalking} through joint modeling of multi-modal inputs (images, speech, and text), enabling dynamic visualization of musical semantics and emotions. As a form of digital performance, singing-capable virtual beings can convey emotional intent with high fidelity, opening new possibilities for immersive music content expression. Recently, Video Diffusion Transformers (DiTs)~\cite{chen2025echomimic, meng2025echomimicv2, jiang2024loopy, tian2024emo, wei2025mocha} have emerged as a unified generative paradigm, widely applied to synthesize highly expressive visual content conditioned on multi-modal signals (e.g., images, speech, and text prompts). Using their powerful spatio-temporal modeling and cross-modal integration capabilities, prior works~\cite{wang2024v, yang2025megactor, qiu2025skyreels, zheng2024memo, lin2025omnihuman, jiang2025omnihuman} have made a significant process in precise facial expression and lip-sync alignment, natural body motion generation, and large-scale data scalability.

However, substantial challenges remain when extending tasks to the complex domain of music performance video generation. First, effective cinematographic language and visual narrative design are critical in music performance. Camera movement patterns, depth-of-field transitions, and compositional beats directly influence audience immersion and emotional delivery. Yet, existing models~\cite{yang2025infinitetalk, tu2025stableavatar, chen2025echomimic, cui2025hallo3, hu2025animateanyone2, ren2025gen3c, kim2025videofrom3d} often lack systematic modeling of cinematographic principles and scene composition, resulting in monotonous framing, rigid motion artifacts, and poor beatic coordination. Current music-driven video generation methods predominantly rely on single-viewpoint or static-scenario configurations, with limited capacity to model multi-camera switching and spatial depth perception—factors crucial for artistic expression and photorealism. Second, cross-modal temporal and beatic alignment remains a key bottleneck. Music-conditioned video content must precisely synchronize camera motions, performance beats, and musical elements (beat, melody, emotion) across temporal dimensions. Ideal systems should not only "understand" music signals and "interpret" textual semantics but also grasp latent emotional intent and narrative logic to generate empathetic, contextually coherent visuals. Moreover, existing methods~\cite{gu2025long, gan2025omniavatar} primarily employ frame-wise sequential prediction, which suffers from content drift and temporal coherence degradation in long-sequence generation. This leads to progressive distortion in identity preservation, pose stability, and expression consistency, ultimately limiting the capacity for high-quality, extended-duration music performance synthesis.

To address these challenges, we introduce YingVideo-MV, a cascaded music-driven video generation framework that integrates music analysis with a temporal-aware diffusion Transformer architecture. This framework enables high-quality talking portrait video generation from music signals while establishing a large-scale Music-in-the-Wild Dataset (MusicMV-Dataset) containing diverse performances. Inspired by the unified perception-action capabilities of intelligent agents~\cite{team2023gemini, xu2025qwen2, ding2025kling, hong2025glm}, we design an MV Director Module that synthesizes multi-modal inputs into structured shot list information. This shot list encapsulates key elements including initial frame composition, beatic cues, and character motion patterns, ensuring alignment between generated content and the intended narrative logic and musical aesthetics. The framework operates through a two-stage pipeline. First, a shot list video is generated from the input music, which then conditions the parallel generation of multiple sub-clips using a music-driven video model. These sub-clips are temporally aligned and visually fused to produce a complete, coherent music performance video with emotional consistency. At the core of our music-driven video model lies a temporal-aware diffusion Transformer that simultaneously handles lip-sync alignment, facial expression generation, and camera motion synthesis, achieving natural coordination among performer actions, musical beat, and camera dynamic. To tackle long-sequence coherence challenges, we propose a dynamic window inference technique that enables seamless transitions and visual consistency in extended video generation by smoothly propagating visual states across adjacent segments. Meanwhile, the incorporation of the DPO training strategy during the optimization process leads to enhanced lip synchronization and improved visual quality.

We summarize our key contributions as follows:
\begin{itemize}
    \item \textbf{MV-Director Framework with Unified instruction Planning.} We propose a global planning module that integrates multi-modal inputs (text prompts, camera trajectories, initial frames, and music segments) into unified semantic instructions. This mechanism elevates music-driven video generation from low-level cue tracking to deep comprehension of musical cinematographic language with explicit control over narrative logic and aesthetic composition.
    \item \textbf{Cascaded Portrait Video Synthesis Pipeline.} We design a two-stage generation framework where first stage establishes high-level semantic and shot planning via MV-Director, while the second stage achieves animation effects within each shot through a temporal-aware diffusion Transformer. This cascaded architecture effectively balances global consistency with local expressiveness, enabling long-sequence video generation with coherent camera movements and smooth transitions, significantly enhancing narrative fluency and musical expressiveness.
    \item \textbf{High-fidelity Generation across diverse scenarios.} YingVideo-MV generates high-fidelity, temporally coherent portrait videos across diverse scenarios. It achieves precise lip-sync alignment, rich facial expression variations, and camera movements synchronized with musical beats. The framework demonstrates superior generalization and artistic controllability, providing a robust foundation for multi-modal content generation applications.
\end{itemize}
Our method produces high-quality music videos with camera movements highly synchronized with the music beat, and highly consistent character portrayals and lip-syncing. YingVideo-MV provides a practical technological approach to automated, high-quality music video creation.

\section{Related Work}

To enable the synchronized generation of music-driven videos with controllable camera trajectories, we explore three research paradigms: audio-driven video generation, controllable camera poses, and joint audio-visual generation of camera trajectories.

\subsection{Audio-driven Video Generation}

This dominant paradigm typically employs video diffusion models for visual content synthesis. Early attempts~\cite{blattmann2023stable, chen2023videocrafter1, zeng2024make} utilized U-Net architectures to demonstrate feasibility, but these models suffered from limited capacity to generate high-fidelity frames with temporal coherence. The advent of Diffusion Transformers~\cite{peebles2023scalable} marked a pivotal advancement, with WAN~\cite{wan2025wan} leveraging their scalable architecture to process spatio-temporal segments at scale, significantly enhancing video coherence and generation quality. Despite these advancements in achieving precise lip synchronization~\cite{chen2025hunyuanvideo, cui2025hallo3, lin2025omnihuman, peng2024synctalk, yariv2024diverse}, two critical limitations persist: (1) weak semantic alignment between audio-visual modalities, and (2) the fixed camera perspective constraint that struggles with occlusion handling and view-dependent deformations under dynamic camera movements. To address these challenges, we introduce an audio-conditioned camera trajectory generation method for rapidly synthesizing animations of camera motion.

\subsection{Controllable Camera Poses}

Precise camera motion control is critical for music MV generation. Early approaches employ fine-tuning techniques such as LoRAs~\cite{hu2022lora} in AnimateDiff~\cite{guo2023animatediff} to manage specific motion types. However, these approaches offer only limited precision. Recent methods like MotionCtrl~\cite{wang2024motionctrl} and its successors~\cite{kuang2024collaborative, xu2024camco, he2024cameractrl} directly condition video generation on extrinsic camera parameters, mapping sparse brush strokes to Gaussian representations. 4D scene generation approaches~\cite{watson2024controlling, wu2025cat4d, sun2024dimensionx} provide inherent camera control through spatio-temporal field modeling, though their synthesis quality currently lags behind specialised video generation models. Motion-I2V~\cite{shi2024motion} and MOFA~\cite{niu2024mofa} introduce two-stage pipelines that first predict motion from strokes then generate videos conditioned on the predicted motion, but this requires maintaining two independent models, increasing system complexity. Recently, TORA~\cite{zhang2025tora} achieved state-of-the-art results by leveraging a DiT backbone with camera-aware positional encodings, while Go-with-the-Flow~\cite{burgert2025go} explores dense trajectory control through optical flow warping mechanisms. However, the above methods still cause camera jumps in the generated video. Therefore, we propose to incorporate camera coding into noise latents over time steps to improve the camera controllability of the generated video.

\subsection{Joint Audio-visual Generation of Camera Trajectories}

While prior approaches focus on deploying video diffusion models trained on millions of in-the-wild videos for controllable generation, a parallel line of research emphasizes capturing high-dimensional volumetric representations through complex data pipelines to enable fine-grained control over video synthesis. By using synchronized multi-camera setups, these methods reconstruct 3D/4D representations (e.g., meshes~\cite{cagniart2010probabilistic, beeler2011high, fyffe2011comprehensive}, NeRFs~\cite{lombardi2019neural, icsik2023humanrf}, or Gaussian splatting~\cite{he2024diffrelight, jiang2024hifi4g, luiten2024dynamic}) for camera trajectory planning, LeviTor~\cite{wang2025levitor} clusters segmentation masks into sparse points enhanced with depth information, but the lack of correspondence modeling and U-Net-based architecture limitations constrain its performance. FlexTraj~\cite{zhang2025flextraj} introduces a multi-granularity, pose-agnostic trajectory control framework for image-to-video generation, enabling flexible camera motion specification without explicit 3D reconstruction. Moreover, all prior approaches assume trajectories are aligned with the first frame, which restricts their applicability. To address these limitations, we propose YingVideo-MV, a cascaded video generation framework, which integrates camera trajectory and audio into DiT backbone.

\section{Method}

The work aims to develop a framework for high fidelity portrait animation generation based on audio supplemented with auxiliary multimodal inputs (including text, static images, and video references). The proposed framework would generate temporally coherent output sequences that maintain the target object's visual identity by the conditional inputs under dynamic camera perspectives, while simultaneously achieving precise alignment with both musical beat patterns and speech-related articulatory features. Key dynamic synthesis attributes of generated video include head movements, gestural dynamics, facial expression variations, and lip-reading consistency synchronised with audio phonemes. In the following sections, we first introduce preliminaries in Sec.~\ref{sec:pre}. Next, we will introduce our cascaded framework and agent-based multimodal operations in Sec.~\ref{sec:pipeline}. We then present our model architecture for music video generation in Sec.~\ref{sec:architec}. As our training strategies,  Direct Preference Optimization (DPO) is applied to align the generated portraits with human aesthetic and perceptual preferences, as detailed in Sec.~\ref{sec:dpo}. Finally, we describe our key strategies for inference in Sec.~\ref{sec:dynamic}.

% During inference, original input video frames are replaced with the sum of random noise and camera pose encodings, as detailed in Sec.~\ref{sec:camera}. A dynamic weighted sliding window denoising strategy is introduced to enhance video smoothness in long-sequence generation by fusing latent information, as detailed in Sec.~\ref{sec:dynamic}. Additionally, Direct Preference Optimization (DPO) is applied to align the generated portraits with human aesthetic and perceptual preferences, as detailed in Sec.~\ref{sec:dpo}

\subsection{Preliminaries}
\label{sec:pre}

\textbf{Flow Matching.} Flow Matching~\cite{lipman2022flow} learns a vector field via conditional flow matching, then transforms the initial noise distribution into target samples through forward ODE integration along stochastic path, entirely bypassing explicit density estimation or inverse transformation procedures. Recent advances~\cite{tan2025ominicontrol, esser2024scaling, wan2025wan, fei2025skyreels} have demonstrated significant performance gains by operating in the latent space of pre-trained autoencoders. Unlike conventional text-to-video models that rely solely on textual conditioning, our approach integrates multi-modal conditioning information comprising driving audio sequences($A$), static portrait images($I$), reference video clips($V$), and associated textual prompts($T$). Crucially, during training, our framework learns to transform random noised camera pose vector sequences ($\mathcal{C}$) into structured and stable dynamic camera trajectory sequences through a decicated denoising progress,

\begin{equation}
    L_{mse}=\mathbb{E}_{z_0, z_1, t\sim [0,1]}[\vert| v_t-u_\theta (z_t, t, T, I, V, A, \mathcal{C}) |\vert ^2_2],
    \label{equ_mse}
\end{equation}

where $u_\theta$ is a trainable denoising net. $z_1$ and $z_0$ notes the latent embedding of the training sample and the initialized noise sampled from the Gaussian distribution $\mathcal{N}(0,1)$. $z_t$ is the training sample constructed using a linear interpolation. Velocity $\mathrm{v}_t=d\mathrm{z}_t/dt=z_1-z_0$ serves as the regression target for the model.

\textbf{Video Diffusion Transformers.} Diffusion Transformers represents a class of generative models built upon the Transformer architecture~\cite{peebles2023scalable}, demonstrating superior performance in video synthesis tasks through the implementation of full spatio-temporal attention across three dimensions. The architecture paradigm enables explicit modeling of long-range dependencies in both spatial and temporal domains, achieving state-of-the-art results in high fidelity video generation. 

\textbf{Low-rank adaptation (LoRA).}
Low-Rank Adaptation (LoRA)~\cite{hu2022lora} is a parameter-efficient fine-tuning strategy. Instead of updating all parameters during fine-tuning, LoRA injects a pair of low-rank matrices into the existing weight layers and restricts optimization to these additional parameters.
By freezing the original weights and training only a small number of low-rank components, LoRA effectively mitigates catastrophic forgetting~\cite{kirkpatrick2017overcoming} while greatly reducing computational overhead. More specifically, the inserted low-rank matrices act as a residual update to the pre-trained weight matrix $\mathbf{W} \in \mathbb{R}^{m \times n}$.
The resulting adapted weight can be formulated as follows:

\begin{equation}
\mathbf{W}^{\prime}=\mathbf{W}+\Delta \mathbf{W}=\mathbf{W}+A B^{T},
\end{equation}

where $A \in \mathbb{R}^{m \times r}$ and $B \in \mathbb{R}^{n \times r}$ denote the two low-rank matrices, and $r$ represents the rank hyperparameter controlling the adaptation capacity. In practical implementations, LoRA modules are typically integrated only into the attention layers of transformer architectures, further decreasing both fine-tuning time and GPU memory consumption.

\textbf{Camera Representation.}
Following prior works~\cite{chen2023ray, kant2024spad, he2024cameractrl, xu2024camco}, we employ the Plücker embedding~\cite{sitzmann2021light} to represent camera poses, as it offers both a strong geometric interpretation and fine-grained per-pixel camera encoding.Specifically, given the camera extrinsic matrix $\mathbf{E} = [\mathbf{R}; \mathbf{t}] \in \mathbb{R}^{3 \times 4}$, where $\mathbf{R}$ and $\mathbf{t}$ denote the rotation and translation components respectively, and the intrinsic matrix $\mathbf{K} \in \mathbb{R}^{3 \times 3}$, we derive the Plücker embedding for each image pixel $(u, v)$ as $\mathbf{p} = (\mathbf{o} \times \mathbf{d}', \mathbf{d}')$.Here, $\mathbf{o}$ represents the camera center in world coordinates, the symbol “$\times$” denotes the cross product, and the ray direction from the camera origin toward the pixel is computed as $\mathbf{d} = \mathbf{R}\mathbf{K}^{-1}[u, v, 1]^T + \mathbf{t}$.
We then normalize $\mathbf{d}$ to obtain $\mathbf{d}'$.
Finally, the Plücker embedding for frame $i$ is expressed as $\mathbf{P}_i \in \mathbb{R}^{6 \times H \times W}$, where $h$ and $w$ correspond to the spatial resolution of the associated visual tokens.

% 使用的backbone 还有backbone的优点，以及相关功能

\subsection{Cascaded Generation Pipeline}
\label{sec:pipeline}

\begin{figure}[t]
    \centering
    \includegraphics[width=\linewidth]{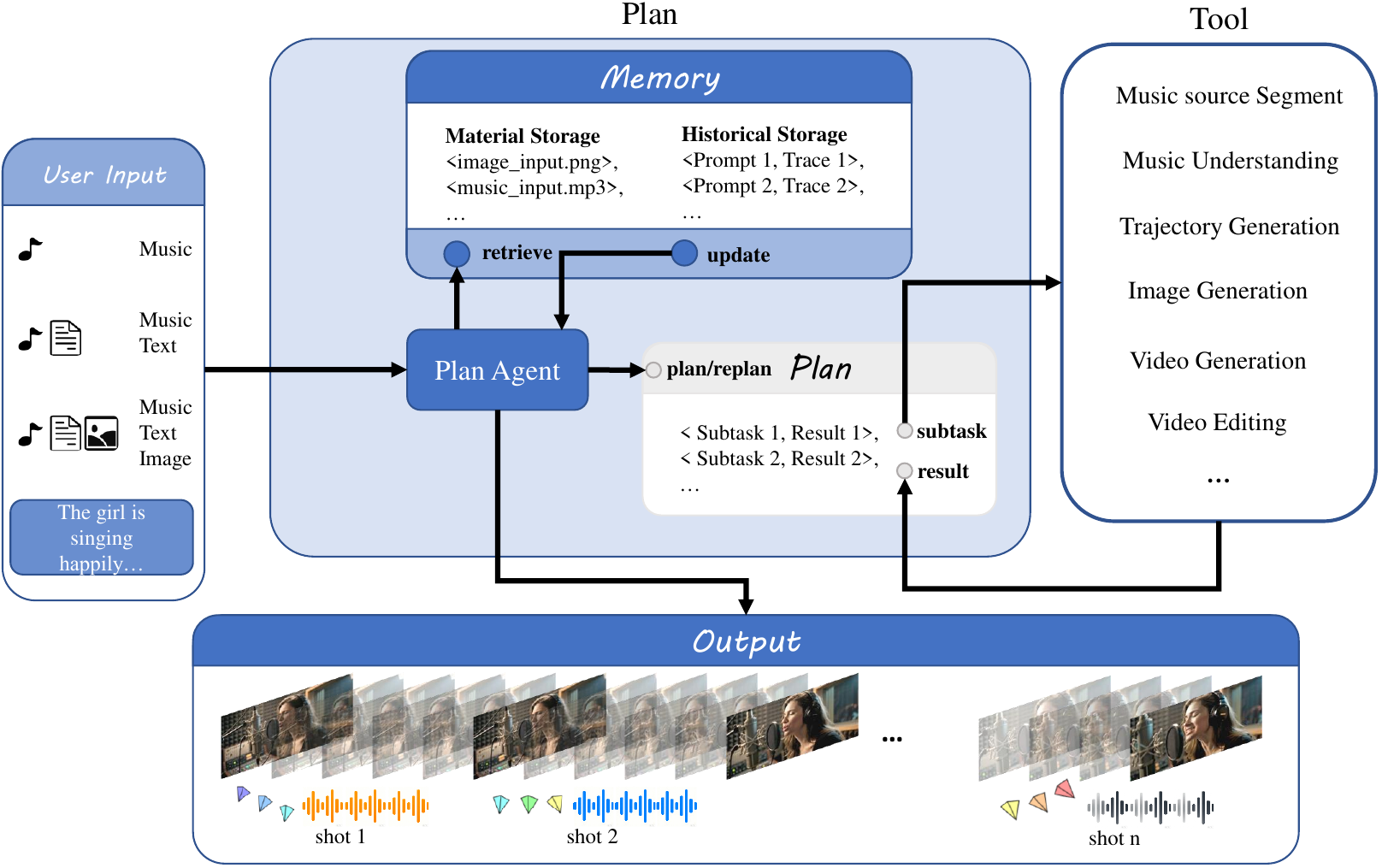}
    \caption{Illustration of YingVideo-MV's cascaded generation Pipeline. Our framework integrates multimodal inputs (music, text, and images) to enable segmented generation of music-performing portrait videos under the guidance of a global planning module. The planning agent strategically invokes specialized tools according to sub-task requirements, ultimately generating three core outputs conditioned on initial-frame specifications: (1) high fidelity music-performing portrait images, (2) coherent dynamic camera trajectories, and (3) synchronized audio sequences aligned with visual performance cues.}
    \label{fig:pipeline}
\end{figure}

As shown in Figure~\ref{fig:pipeline}, we formulate the task of multimodal music video (MV), creation as a sequential decision-making problem. The MV-Director agent operates in an environment defined by a user-specified high-level goal $G$ — such as narrative intent, visual style , emotional tone, or character design — and a toolbox $T$ containing various multimodal operations, including music analysis, scene planning, image/video generation, editing, and refinement modules. The agent aims to synthesize a sequence of actions $A = (a_1, a_2, \ldots, a_N)$ that transforms the initial state $s_0$ (consisting of music, user prompts, and optional reference images or videos) into a final state $s_N$ that satisfies the MV creation goal $G$. The core challenge is one fold — designing a rich and accurate toolbox $T$ that covers the full spectrum of MV production needs.

\textbf{Toolbox $T$.} The toolbox $T$ includes music source segmentation, music understanding, trajectory generation, image generation, video generation, and video editing. These tools allow the agent to analyze music, design scenes, synthesize visuals, and iteratively refine the video. 

\textbf{Music source segmentation}. To segment raw music into semantically meaningful clips, we iteratively identify local maxima in the onset strength of beats as potential boundaries, ensuring that the resulting average segment duration approximates one musical bar:
\begin{equation}
    \Delta_{bar} = 4\Delta_{beat}=4\cdot\frac{60}{\textrm{bpm}}.
\end{equation}
The segmentation satisfies two principles: (1) According to the \textit{cut-to-the-beat} rule, secne boundaries should conincide with strong beats; (2) The duration of MV scenes typically correlates with the bar length~\cite{Laure2021isthere}. 

\textbf{Music understanding.} Inspired by multimodal large language models (MLLMs)~\cite{bai2025qwen2, hong2025glm, qi2025lmm}, we unify music-related evidence into a shared semantic space to provide high-level control signals for global MV planning. Specifically, we employ Qwen 2.5-Omni~\cite{xu2025qwen2} to extract both the transcription and the emotional attributes from each music segment, enabling the system to generate scene-level script content that matches the musical style, mood, and dynamics. 

\textbf{Trajectory generation}. Camera trajectory design is critical for MV production, as it directly influences visual storytelling, shot composition and emotional pacing. Existing methods often rely on geometric heuristics or learning-based approaches that lack textual alignment or fine-grained control. Following recent advances in cinematography modeling (e.g., GenDoP~\cite{zhang2025gendop}), our trajectory generator leverages scene content and script-level guidance to produce smooth, context-aware camera movements. This module generates expressive and musically aligned camera paths, supporting shot framing, depth transitions, and narrative continuity throughout the MV.

\textbf{Image generation}, \textbf{video generation}, and \textbf{video editing}. This module handles the creation and refinement of visual assets for each scene. Image generation is supported by open-source models such as Flux and SDXL, or closed-source APIs like MidJourney for text-to-image synthesis and reference-guided consistency. Video generation converts images and trajectories into temporally coherent clips, leveraging the S2V model to produce smooth sequences aligned with musical beats. Finally, video editing performs post-processing, including video and music segment concatenation, subtitle addition, and overall stylistic refinement ensuring narrative and visual consistency across the entire music video.

% our cascaded generation pipeline draws inspiration from Kling-Avatar to achieve intent-aware video synthesis. To enable comprehensive understanding of multi-modal inputs, we project audio, image, and text signals into a shared semantic space, generating high-level control signals for global planning. Specifically, we first extract audio captions from the input audio using a pre-trained speech-to-text model and generate image captions from the reference portrait via a vision-language model. These captions, combined with user-provided text prompts, are processed by the MV-Director module to establish semantic correlations across memory, audio, and visual cues. The module organizes these elements into a unified textual instruction, which is injected into the video diffusion model through cross-attention layers during generation. This hierachical approach ensures that the final music performance video aligns with both the explicit input conditions and the implicit artistic intent behind the instructions.

\subsection{Model Architecture}
\label{sec:architec}

\begin{figure}
    \centering
    \includegraphics[width=\linewidth]{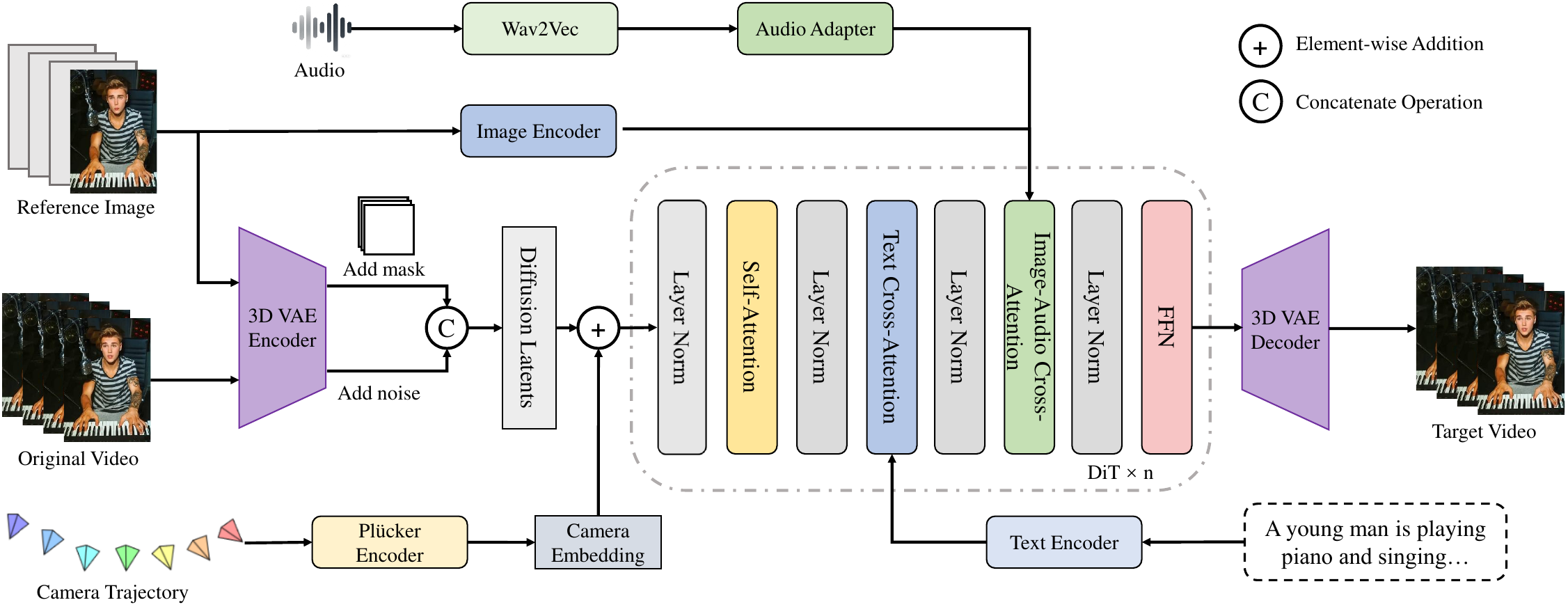}
    \caption{Illustration of Video Generation Model Architecture. Embeddings from the image and text encoders are injected into each block of the DiT. Given music audio input, we leverage Wav2Vec to extract audio embeddings, while the camera trajectory is encoded and incorporated into the diffusion latent. To model the joint audio–latent representation, audio embeddings are fed into an audio adapter, the outputs of which are injected into the DiT via cross-attention.}
    \label{fig:architect}
\end{figure}

As illustrated in Figure~\ref{fig:architect}, the S2V module of YingVideo-MV builds upon the widely adopted WAN 2.1~\cite{wan2025wan} framework and follows established research paradigms. Music audio inputs are first processed through Wav2Vec to extract audio embeddings, which are subsequently refined using StableAvatar~\cite{tu2025stableavatar}'s audio adapter to mitigate potential distribution mismatches. These enhanced embeddings are then fed into the denoising DiT pipeline. Reference images are processed via two parallel pathways: (1) Temporal axis padding with zero-filled frame is followed by latent encoding through a frozen 3D VAE encoder. The resulting latent codes are concatenated along the channel dimension with compressed video frames and binary masks (1 for the first frame, 0 otherwise); (2) Image embeddings are generated via a CLIP image encoder and injected into every image-audio cross-attention block of the denoising DiT to regulate visual appearance.

\textbf{Camera controlled video generation}. After obtaining the Plücker embedding~\cite{sitzmann2021light} $\mathbf{P}_i$ that encodes the camera pose of the $i$-th frame, we represent the complete camera trajectory of a video clip as a sequence of Plücker embeddings $\mathbf{P} \in \mathbb{R}^{L \times 6 \times H \times W}$, where $L$ denotes the video clip length. To inject camera information into the video generation backbone and enable explicit camera control, inspired by previous works~\cite{he2024cameractrl, bai2025recammaster}, we employ an adapter module to project the camera embeddings so that their tensor shape matches that of the noisy latent. The projected embeddings are then fused with the latent via element-wise addition before being fed into the DiT model. This design provides a simple yet effective mechanism for conditioning video generation on camera motion. The camera adapter architecture is composed of a sequence of \texttt{PixelUnshuffle}, \texttt{Conv2d}, and \texttt{ResidualBlock} layers.

\subsection{Direct Preference Optimization}
\label{sec:dpo}

\begin{figure}[t]
    \centering
    \small
    \includegraphics[scale=0.8]{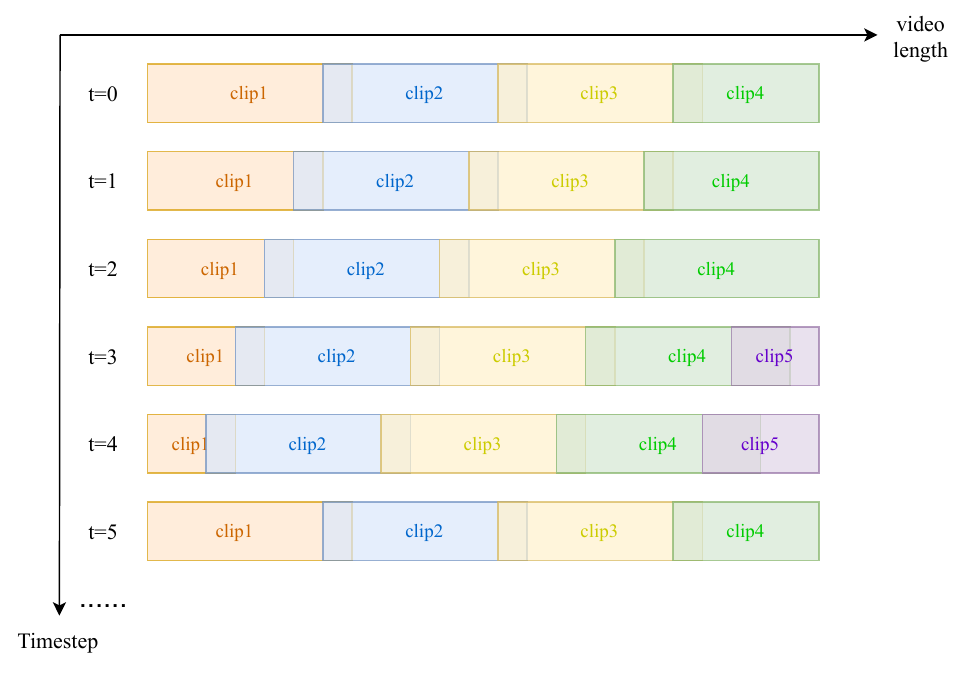}
    \caption{Timestep-aware dynamic window range strategy. From top to bottom, each row represents a denoising process at one timestep. Within each row, each clip of different color represents the segmentation of the long video. There are overlapping areas between each clip. At t=3, the last clip expands its overlap with the preceding clip to satisfy the minimum clip-length constraint. At t=5, the starting offset is reset because the offset accumulated in the previous timestep has reached its maximum allowable value.}
    \label{fig:time-aware}
\end{figure}

Direct Preference Optimization (DPO) formalizes the alignment of the model with human preferences as a policy optimization task based on pairwise preference data. Let the preference dataset be denoted as $D = \{(x, y^w, y^l)\}$, where $y^w$ represents the preferred sample and $y^l$ represents the less-preferred one. For each training sample, we randomly select four video clips and compute three evaluation metrics for each clip: the Sync-C score~\cite{li2024latentsync}, the hand-quality reward score (Hand-Specific Reward~\cite{cui2025hallo4}), and the VideoReward score~\cite{liu2025improving}. After aggregating these three indicators through weighted scoring, the segment with the highest composite score is designated as $y^w$, while the one with the lowest score is designated as $y^l$.

The DPO objective aims to maximize the likelihood of preferred outputs while regularizing the deviation from a reference policy $\pi_{\mathrm{ref}}$. Formally, it is defined as:
\begin{equation}
    \mathcal{L}_{\mathrm{DPO}}=-\mathbb{E}_{\left(x, y^{w}, y^{l}\right) \sim \mathcal{D}}\left[\log \sigma\left(\beta \log \frac{\pi_{\theta}\left(y^{w} \mid x\right)}{\pi_{\mathrm{ref}}\left(y^{w} \mid x\right)}-\beta \log \frac{\pi_{\theta}\left(y^{l} \mid x\right)}{\pi_{\mathrm{ref}}\left(y^{l} \mid x\right)}\right)\right], 
\end{equation}
where $\sigma(\cdot)$denotes the sigmoid function, and $\beta$ is a temperature coefficient controlling the strength of regularization towards $\pi_{\text{ref}}$.

Following the Flow-DPO loss proposed in VideoReward, the DPO loss in our training process can be reformulated as:
\begin{equation}
    \begin{aligned}
\mathcal{L}_{\mathrm{DPO}}=-\mathbb{E}_{\left(y^{w}, y^{l},t\right) \sim \mathcal{D}}\left[\log\sigma\left(-\frac{\beta_{t}}{2}\left(\left\|\boldsymbol{v}^{w}-\boldsymbol{v}_{\theta}\left(\boldsymbol{x}_{t}^{w}, t\right)\right\|^{2}-\left\|\boldsymbol{v}^{w}-\boldsymbol{v}_{\mathrm{ref}}\left(\boldsymbol{x}_{t}^{w}, t\right)\right\|^{2}\right.\right.\right. \\
\left.\left.\left.-\left(\left\|\boldsymbol{v}^{l}-\boldsymbol{v}_{\theta}\left(\boldsymbol{x}_{t}^{l}, t\right)\right\|^{2}-\left\|\boldsymbol{v}^{l}-\boldsymbol{v}_{\mathrm{ref}}\left(\boldsymbol{x}_{t}^{l}, t\right)\right\|^{2}\right)\right)\right)\right]
\end{aligned}
\end{equation}

Here, $v_{ref}$ refers to the velocity field of the reference model, initialized from the diffusion model, while $v^w$ and $v^l$ correspond to the velocity fields obtained from the preferred sample $y^w$ and the dispreferred sample $y^l$. During training, gradients from $L_{DPO}$ adjust the denoising direction to favor high-reward regions in the trajectory space while preserving the temporal dynamics of the reference policy. This joint optimization enables preference-aware generation without compromising the inherent stability of pretrained diffusion models.

\subsection{Timestep-aware Dynamic Window Range Strategy}
\label{sec:dynamic}

\begin{algorithm}
\caption{Timestep-aware dynamic window range strategy}
\label{alg:time-aware}
\begin{algorithmic}[1]

\Ensure Video generator $\mathrm{G}(\cdot)$ with window length $f$, 
audio $c_a^{[0,l]}$, camera $c_c^{[0,l]}$, text $c_t$, image $c_i$, steps $T$.
\Require Final denoised latent $z_0^{[0,l]}$.

\State Initialize noisy latent $z_T^{[0,l]}$, shift offset $\alpha$, shift step $p$, max offset $m$, min clip length $n$, overlap length $o$
\For{$t = T, \ldots, 1$}

    \If{$\alpha > m$}
        \State $\alpha \gets 0$ \Comment{reached max offset, reset shift}
    \EndIf

    \State $s \gets -\alpha$
    \State $e \gets s + f$
    \State $s \gets \max(0, s)$ \Comment{ensure no rolling}

    \While{$e < l$}
        \State $z_{t-1}^{[s,e]} \gets 
            \mathrm{G}(z_t^{[s,e]}, c_a^{[s,e]}, c_c^{[s,e]}, c_t, c_i, t)$

        \If{$e < l$}
            \State $s \gets e - o$
            \If{$s + f < l$}
                \State $e \gets s + f$
            \Else
                \State $e \gets l$
                \If{$e - s < n$}
                    \State $s \gets e - n$ \Comment{ensure minimum clip length}
                \EndIf
            \EndIf
        \EndIf

    \EndWhile

    \State $\alpha \gets \alpha + p$ \Comment{accumulate shift}

\EndFor

\end{algorithmic}
\end{algorithm}

% TODO：修改优化 

For long-video generation strategies, traditional approaches first denoise entirely one video clip latent, then moving on to the next clip. To maintain temporal coherence, the last few frames of the previous clip are used as motion frames to provide conditioning for the subsequent clip. Differently, recent works (e.g., Sonic ~\cite{ji2025sonic} and Stable Avatar ~\cite{tu2025stableavatar}) process the entire sequence of latent clips at each diffusion timestep, then move to next timestep. Additionally, Sonic shifts the starting position at every timestep to enlarge the contextual receptive field of each frame, but its clips have no overlap. In contrast, Stable Avatar restarts denoising from the beginning at every timestep without shifting the starting position, but its clips have overlap.

Unlike Sonic, we do not employ a rolling strategy; that is, the initial frames and the final frames are not merged into a single clip for denoising, as these two portions of the video may have undergone substantial visual changes and are therefore unsuitable for self-attention. And unlike Stable Avatar, our method does shift the starting position at each timestep.

Furthermore, our Timestep-aware dynamic window range strategy pays special attention to the first and last clips. After several steps of shifting, the number of frames in the first clip may become very small. We observe that denoising within such a small window degrades generation quality, so we introduce a constraint: once the first-clip length shrinks to a threshold, the next timestep resets to zero starting offset like timestep 0. Similarly, the last clip may also become too short. Our solution is to extend it forward until it reaches the minimum clip length requirement, which increases the number of overlapping frames between the last and the second-last clips.

As illustrated in Figure~\ref{fig:time-aware} and Algorithm~\ref{alg:time-aware}, our Timestep-aware dynamic window range strategy consists of two nested loops. The outer loop is inverse diffusion process, while the inner loop is sliding window process that model predicts for the each clip on the audio conditions. Within the inner loop, our strategy divides the video into different clips, the position and length of each clip dynamically change at each time step. 

\section{Experiments}

\textbf{Implementation Details.} We adopt the Wan2.1-I2V-14B architecture as the baseline video diffusion model for our experiments, which was trained using a constant learning rate of $1 \times 10^{-5}$. The MV-Director is powered by a finetuned Qwen 2.5-Omni~\cite{xu2025qwen2} model. The proposed framework was trained on 64 NVIDIA A800-80G GPUs with mixed-precision acceleration. For stage-1 training, we curated a general-purpose video dataset containing approximately 1,500 hours of single-person facial/body performances, with individual clips averaging 10 seconds in duration. To enhance musical performance expressiveness in stage-2 training, we further incorporated 400 hours of domain-specific music performance videos featuring synchronized audio-visual recording of professional singers and virtual avatars. % All training data were rigorously preprocessed, including face detection, audio-text alignment vertification, and quality filtering to ensure high-fidelity generation capabilities.

\textbf{Test Datasets and Evaluation Metrics.} Following established benchmarks, we evaluate our method’s performance across diverse scenarios using three datasets (HDTF~\cite{zhang2021flow}, CelebV-HQ~\cite{zhu2022celebvhq}, and EMTD~\cite{meng2025echomimicv2}) and a a camera motion dataset (MultiCamVideo). Performance is quantified through complementary automated metrics and human evaluation. For objective assessment, we employ the following: Fréchet Inception Distance (FID) to measure per-frame visual quality; Fréchet Video Distance (FVD) for temporal consistency evaluation; SyncNet-based Sync-C (confidence score) and Sync-D (lip distance) to qualify lip-sync accuracy; Cosine Similarity (CSIM) scoring identity preservation; and rotation error (RotErr) / translation error (TransErr)~\cite{he2024cameractrl} to assess camera motion precision. To capture perceptual nuances beyond automated metrics, we conduct user studies where 20 participants rate 15 generated videos from five dimensions; gesture synchronization with music audio prosody, body motion alignment with speech beat, lip-sync accuracy, identity consistency, and overall naturalness, yielding 300 feedback samples for comparative analysis.

\subsection{Experiment Results}

\begin{figure}
    \centering
    \includegraphics[width=\linewidth]{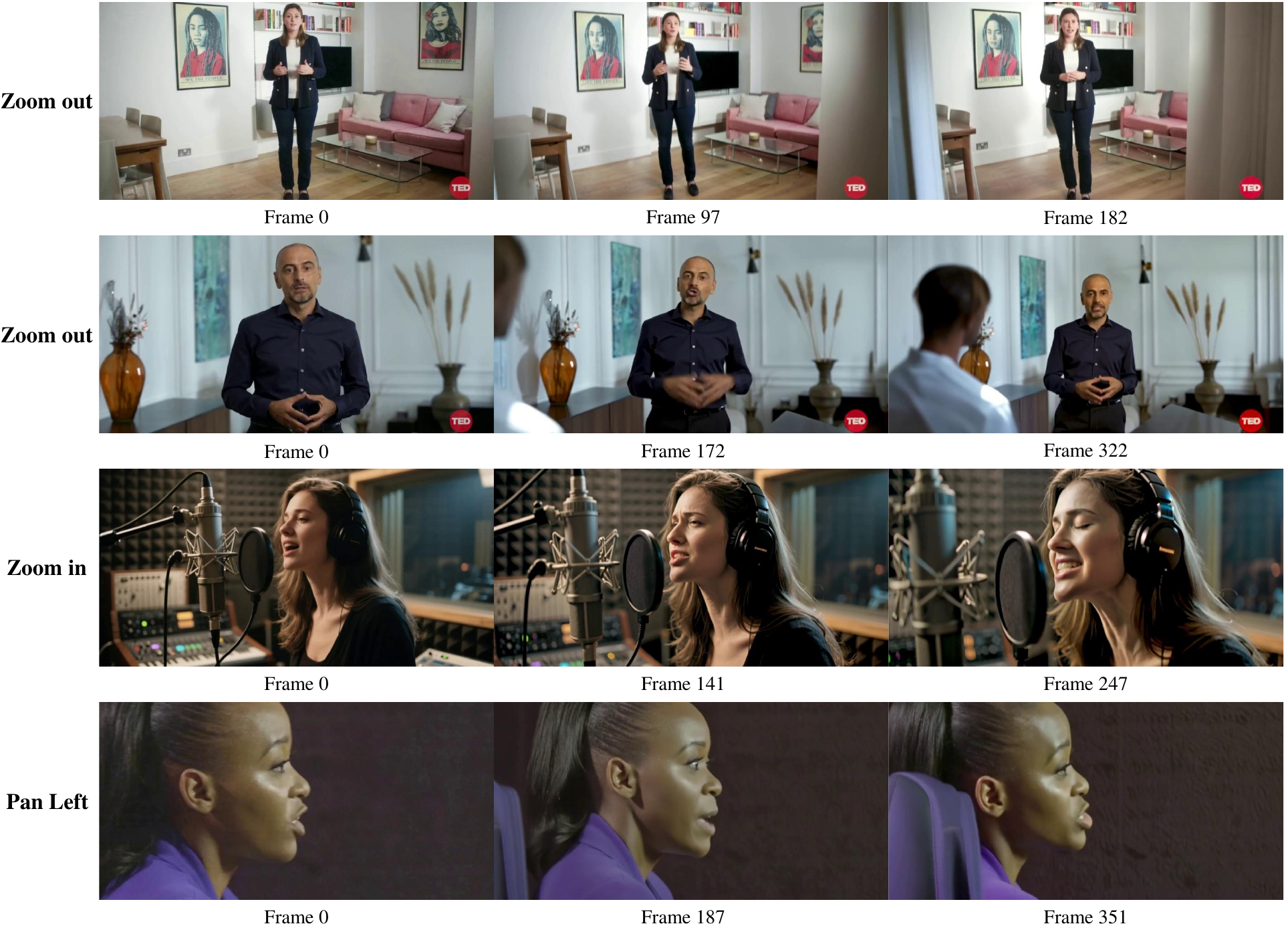}
    \caption{\textbf{Visualization of Camera Movement.} This figure illustrates the music-driven high performance of our framework with synchronized camera motions. The generated sequences demonstrate precise alignment between body movements and camera motion.}
    \label{fig:cam_mov}
\end{figure}

\textbf{Qualitative Results}. Figure~\ref{fig:cam_mov} presents music-driven long-video generation results under various camera controls. Our method demonstrates superior performance in identity preservation, motion clarity, and expression control. For instance, in the first two rows, our method exhibits both structural coherence and seamless transitions during scene expansion under camera zoom-out, effectively maintaining spatial consistency while generating plausible out-of-frame content. In the third row, even under camera zoom-in conditions, YingVideo-MV preserves character identity fidelity and fine-grained scene details (e.g., background textures and lighting consistency). Notably, the generated sequences balance cinematographic precision with visual naturalness across all evaluated camera motion patterns, including complex transitions between dynamic and static shots.

\textbf{Quantitative Results}. As shown in Table~\ref{tab:comparison}, we compare our method against camera motion-aware models (CameraCtrl~\cite{he2024cameractrl}, Uni3C~\cite{cao2025uni3c}) and long-sequence generation frameworks (StableAvatar~\cite{tu2025stableavatar}, InfiniteTalk~\cite{yang2025infinitetalk}). Due to the absence of camera motion capabilities in StableAvatar and InfiniteTalk, their performance on camera-related metrics cannot be evaluated. Our method demonstrates significantly better visual quality and identity preservation, outperforming InfiniteTalk's sparse frame audio-driven method (which achieves good FID and FVD scores through local lip region optimization, while our method suffers from reduced FID and FVD scores due to camera movement). While our method does not achieve the minimal rotation error, it surpasses other baselines in translation error, demonstrating more natural camera trajectory estimation. The balanced lip-sync accuracy and camera motion generation indicates a harmonious integration of multimodal temporal alignment and cinematographic control, representing a significant advancement in music-driven video synthesis.

\begin{table}[ht]
    \centering
    \resizebox{\textwidth}{!}{
    \begin{tabular}{c c c c c c c c c c}
        \multirow{2}{*}{Method} & \multicolumn{2}{c}{Function} & \multicolumn{7}{c}{Metrics} \\
        \cline{2-3} \cline{4-10}
        &\makecell[c]{Camera \\ movement} & \makecell[c]{Long video\\ generation} & RotErr $\downarrow$ & TransErr $\downarrow$ & FID $\downarrow$ & FVD $\downarrow$ & CSIM $\uparrow$ & Sync-C $\uparrow$ & Sync-D $\downarrow$ \\
        \midrule
        Stable Avatar~\cite{tu2025stableavatar} & \XSolidBrush & \Checkmark & - & - & 38.14 & 375 & 0.635 & 5.15 & 9.49 \\
        InfiniteTalk~\cite{yang2025infinitetalk} & \XSolidBrush & \Checkmark & - & - & \textbf{27.14} & \textbf{132.54} & \underline{0.744} & \underline{5.61} & \underline{9.18} \\
        CameraCtrl~\cite{he2024cameractrl} & \Checkmark & \XSolidBrush & \underline{1.18} & 9.02 & 36.72 & 360.3 & 0.498 & 4.35 & 11.29 \\
        Uni3C~\cite{cao2025uni3c} & \Checkmark & \XSolidBrush & \textbf{1.13} & \underline{7.26} & 31.75 & 253.76 & 0.574 & 4.66 & 10.03 \\
        Ours & \Checkmark & \Checkmark & 1.22 & \textbf{4.85} & \underline{30.36} & \underline{193.68} & \textbf{0.753} & \textbf{6.07} & \textbf{8.67} \\
        \bottomrule
    \end{tabular}
    }
    \caption{\textbf{Quantitative comparison with other methods.} This table summarizes common camera motion and long-sequence video generation methods, with our approach uniquely combining both capabilities. Quantitative metrics demonstrate superior performance where \textbf{Bold} indicates the best result and \underline{Underline} denotes the second-best across all evaluated benchmarks.}
    \label{tab:comparison}
\end{table}

\textbf{User Study}. To further validate our method's effectiveness, we conducted a subjective evaluation on our internal dataset, which is shown in Table~\ref{tab:user_study}. Participants rated four key dimensions using a 5-point scale with 0.5 increments (1=worst, 5=best): (1) smoothness and coherence of camera motions, (2) lip-sync accuracy with musical beats, (3) naturalness of character movements, and (4) overall video quality. The results demonstrated statistically significant superiority of our method across all dimensions, with average scores of 4.3$\pm$0.6 for camera motion, 4.5$\pm$0.5 for lip-sync, 4.2$\pm$0.5 for motion naturalness, and 4.4$\pm$0.6 for overall quality. Inter-rater reliability analysis showed substantial agreement, confirming the robustness of subjective assessments. Qualitative feedback highlighted our framework's ability to generate cinematographically coherent sequences with natural audio-visual synchronization, particularly in complex scenarios involving dynamic camera transitions and rapid lip movements.

\begin{table}[ht]
    \centering
    \resizebox{\textwidth}{!}{
    \begin{tabular}{c c c c c}
        Method & \makecell[c]{Smoothness and coherence\\ of camera motions} & \makecell[c]{Lip-sync accuracy\\ with musical beats} & \makecell[c]{Naturalness of\\ character movements} & Overall video quality \\
        \midrule
        Stable Avatar~\cite{tu2025stableavatar} & 1.3$\pm$0.2 & 3.9$\pm$0.4 & 3.7$\pm$0.3 & 3.9$\pm$0.2 \\
        InfiniteTalk~\cite{yang2025infinitetalk} & 1.4$\pm$0.4 & \underline{4.4$\pm$0.4} & \textbf{4.3$\pm$0.3} & \underline{4.4$\pm$0.4} \\
        CameraCtrl~\cite{he2024cameractrl} & 3.8$\pm$0.3 & 3.7$\pm$0.1 & 3.4$\pm$0.2 & 3.3$\pm$0.3 \\
        Uni3C~\cite{cao2025uni3c} & \underline{4.0$\pm$0.1} & 4.0$\pm$0.7 & 3.6$\pm$0.2 & 3.7$\pm$0.3 \\
        Ours & \textbf{4.3$\pm$0.6} & \textbf{4.5$\pm$0.5} & \underline{4.2$\pm$0.5} & \textbf{4.4$\pm$0.6} \\
        \bottomrule
    \end{tabular}
    }
    \caption{\textbf{User Study.} The metrics demonstrate superior performance where \textbf{Bold} indicates the best result and \underline{Underline} denotes the second-best across all evaluated benchmarks.}
    \label{tab:user_study}
\end{table}

\begin{figure}
    \centering
    \includegraphics[width=\linewidth]{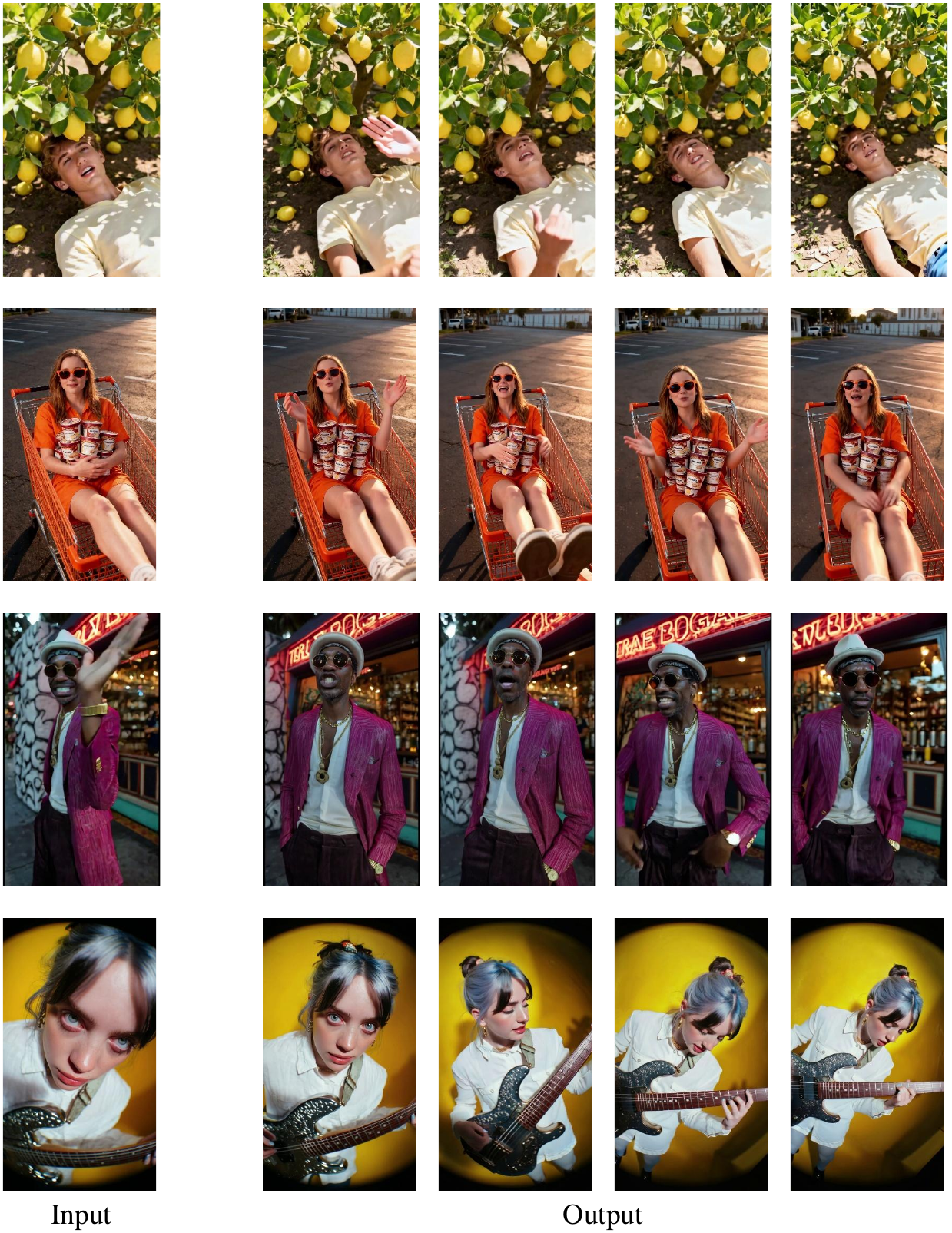}
    \caption{\textbf{More generated results of YingVideo-MV}. The figure shows the video results generated by YingVideo-MV using various camera motion combinations.}
    \label{fig:more}
\end{figure}

\subsection{Ablation Study}

We conduct ablation experiments to quantitatively evaluate the impact of reward feedback mechanisms and temporal coherence strategies. Specifically, we train and test models with and without DPO optimization, as well as with and without the dynamic window inference strategy. As shown in the Table~\ref{tab:ablation}, the introduction of DPO leads to improvements across all metrics, demonstrating enhanced video quality, lip-sync accuracy, and identity preservation. Similarly, the dynamic window strategy achieves 6.3\% improvement in temporal smoothness (FVD$\downarrow$). These results indicate that our combination of DPO-driven preference alignment and dynamic window-based temporal coherence not only improves technical fidelity and synchronization but also enhances expressiveness, producing outputs that better align with human perceptual preferences.

\begin{table}[ht]
    \centering
    \begin{tabular}{c c c c c c}
         & FID $\downarrow$ & FVD $\downarrow$ & CSIM $\uparrow$ & Sync-C $\uparrow$ & Sync-D $\downarrow$  \\
        \midrule
        Ours & \textbf{30.36} & \textbf{193.68} & \textbf{0.753} & \textbf{6.07} & \textbf{8.67} \\
        Ours(w/o DPO) & 35.02 & 203.71 & 0.728 & 5.88 & 8.92 \\
        Ours(w/o TDW) & 35.63 & 205.88 & 0.731 & 5.79 & 9.03 \\
        \bottomrule
    \end{tabular}
    \caption{\textbf{Quantitative results of ablation study}. 'TDW' means timestep-aware dynamic window range strategy.}
    \label{tab:ablation}
\end{table}

\section{Limitation and Future Work}

Our framework currently faces challenges in generating animations for non-human entities when provided with reference images exhibiting significant morphological and structural differences from human subjects. While our method achieves high performance on human avatars, the geometric and textural complexities of fantastical creatures (e.g., multi-limbed beings, non-biological structures) exceed the current model's capacity for novel shape synthesis. A promising direction involves integrating an auxiliary reference-aware network to explicitly capture semantic details through hierarchical feature adaptation and cross-domain knowledge transfer. We also observe limitations in modeling complex interpersonal dynamics - extending our framework to support multi-character interactive music videos (MC-MV) remains a key objective for future exploration. This would require advancing interpersonal spatial reasoning and behavioral coordination modeling while maintaining strict audio-visual synchronization across multiple agents.

\section{Conclusion}

In this work, we introduced YingVideo-MV, a cascaded video generation framework that unifies multi-modal inputs for long-sequence music video synthesis. Our two-stage pipeline first employs MV-Director for global shot planning, followed by DiT-based clip-wise generation to produce high-resolution video details. A dynamic window inference optimization module further refines inter-clip visual transitions to address the critical challenges of cinematic language deficiency and temporal inconsistency in music video generation. Combined with meticulously curated datasets and practical training/inference strategies, our framework achieves faithful global semantic alignment while preserving fine-grained audio-visual details. Experimental results demonstrate YingVideo-MV's ability to generate videos with precise lip synchronization, identity consistency, and controllable camera dynamics. Superiority over baseline methods is further validated through human preference-based metrics. We believe that this work provides practical solutions for automated, high-quality music content creation and establishes a novel paradigm for future multi-modal video generation systems.

% \section{Author}

% Jiahui Chen, Weida Wang, Runhua Shi, Huan Yang, Chaofan Ding, Zihao Chen.

\bibliographystyle{rusnat}
\bibliography{bib}


\end{document}